\documentclass[10pt, a4paper]{article}

\usepackage[]{lrec-coling2024} 

\usepackage{xurl}
\usepackage{lscape}
\usepackage{numprint}
\usepackage{subfig}

\title{M2SA: Multimodal and Multilingual Model for Sentiment Analysis of Tweets}
\name{Gaurish Thakkar$^{1}$, Sherzod Hakimov$^{2}$, Marko Tadić$^{1}$}

\address{$^1$Faculty of Humanities and Social Sciences, University of Zagreb\\ $^2$Computational Linguistics, University of Potsdam \\
         \{gthakkar, marko.tadic\}@ffzg.hr, 
         first.last@uni-potsdam.de\\
         }

\abstract{
In recent years, multimodal natural language processing, aimed at learning from diverse data types, has garnered significant attention. However, there needs to be more clarity when it comes to analysing multimodal tasks in multi-lingual contexts. While prior studies on sentiment analysis of tweets have predominantly focused on the English language, this paper addresses this gap by transforming an existing textual Twitter sentiment dataset into a multimodal format through a straightforward curation process. Our work opens up new avenues for sentiment-related research within the research community. Additionally, we conduct baseline experiments utilising this augmented dataset and report the findings. Notably, our evaluations reveal that when comparing unimodal and multimodal configurations, using a sentiment-tuned large language model as a text encoder performs exceptionally well. \\ \newline \Keywords{sentiment analysis, multilingual, multimodal} 
}
\begin{document}

\maketitleabstract
\section{Introduction}


Social media platforms serve as conduits for the dissemination of information. Tweets have emerged as a trendy medium through which individuals communicate and express their ideas and opinions. Twitter (aka \textit{X}) is widely used by researchers as a prominent social media platform for engaging in micro-blogging and fostering interactions. Sentiment analysis \citep{10.3115/1219840.1219855} is a well-studied topic in natural language processing. The topic has received consideration in both unimodal and multimodal contexts. The proliferation of social media platforms, including Twitter and YouTube, has led to a common practice of assessing content using several modalities~\citep{you2016robust, yu-etal-2020-ch}. This approach offers additional context through spoken, non-verbal, and auditory aspects. The primary focus in many domains of natural language processing (NLP) often revolves around higher-resourced languages. However, the challenge of processing lower-resourced languages remains unresolved.

The process of annotating supervised datasets for natural language processing (NLP) tasks is a labour-intensive endeavour requiring significant investment of time, financial resources, and effort. Recently, several shared tasks, including SemEval~\citep{nakov-etal-2016-semeval, ghosh-etal-2015-semeval}, have introduced tasks aimed at identifying the polarity of tweets, categorising them into predetermined classes. All of the datasets for the shared tasks are accompanied by labels considered the gold standard. Another point to take into account here is that previous approaches~\citep{Raffel2019ExploringTL,xie2020unsupervised,cliche2017bb_twtr} focused on text-only, while posts shared on social media sometimes include images, videos, etc. Approaches incorporating multimodal information~\citep{7837868, DBLP:conf/mir/CheemaHME21, PORIA201650} for the classification of sentiment are predominantly focused on the English language.

This paper presents a straightforward approach for enhancing pre-existing publicly accessible datasets to conduct multimodal (image \& text) sentiment analysis on Twitter called \textit{M2SA} (Multimodal Multilingual Sentiment Analysis). We have collected existing datasets in 21 languages where each annotated post includes both text and image with the annotated labels being either \textit{positive}, \textit{negative}, or \textit{neutral}. We then trained a multimodal model that combines image and text embedding features to classify the target labels.

Our contributions are as follows:
\begin{itemize}
    \item We engage in curating, enriching, and analysing pre-existing Twitter sentiment datasets in 21 different languages. 
    \item The pre-trained model architectures use a fusion of textual information and visual features, utilising large language models for text encoding and image encoding.
    \item The study examines the effects of utilising machine translation instances in the context of lower-resourced languages.
\end{itemize}

All resources (pre-trained models, datasets) and the source code are shared publicly\footnote{\url{https://github.com/cleopatra-itn/M2SA-multimodal-multilingual-sentiment-analysis}}. The subsequent sections of the paper are structured in the following manner: Section 2 provides an overview of the existing literature and research in the field. Section 3 provides a comprehensive overview of the processes involved in data collection, enrichment, and the statistical characteristics of the dataset. The methodology for classification is outlined in Section 4. The experimental setup and results are outlined in Sections 5 and 6. The paper concludes in Section 7.

\section{Related Work}
\citet{7837868} presented a framework that uses CNNs to extract features from multimodal data's visual and textual modalities. The visual features are extracted using a CNN model that has been previously trained, such as VGG16 or ResNet-50. A CNN model trained on a massive corpus of text data is used to extract the textual features. The combined extracted features from the visual and textual modalities are then fed into an MKL classifier. The MKL classifier discovers the optimal combination of kernels for distinguishing between distinct emotions or sentiments.
\citet{PORIA201650} used both feature and decision-level fusion methods to merge affective information extracted from multiple modalities. \citet{DBLP:conf/mir/CheemaHME21} evaluated various embedding features from both text and visual content. \citet{huang2023multimodal} proposed a new framework for multimodal sentiment analysis in realistic environments, with two main components: a module for multimodal word refinement and a module for cross-modal hierarchical fusion. \citet{baecchi2016multimodal} employed a strategy that uses a skip-gram neural network to extract features from the text mode. Image-specific features are extracted using a denoising autoencoder \citep{10.5555/1756006.1953039} neural network. The denoising autoencoder network is taught to reconstruct an image from its corrupted version. The extracted features from the text and image modalities are then concatenated and fed to an SVM classifier. In addition to considering the modelling strategies for the sentiment analysis task, it is essential to identify the available benchmarking datasets. English contains a substantial quantity of multimodal datasets on sentiment and emotion analysis \citep{Go2009, Mohammad2018}. While the TweetEval \citep{barbieri2020tweeteval} examines the application of large language models to seven tasks in Twitter, including emotion, emoji, irony sentiment, and others, the test set is monolingual. The paper authored by \citet{garg-etal-2022-multimodality} provides a comprehensive exposition on diverse multimodal datasets, encompassing the domain of multimodal sentiment analysis.

\section{Multimodal Multilingual Sentiment Analysis (M2SA)}

\begin{figure*}[!ht]
\begin{center}
\centering
\includegraphics[width=0.8\textwidth]{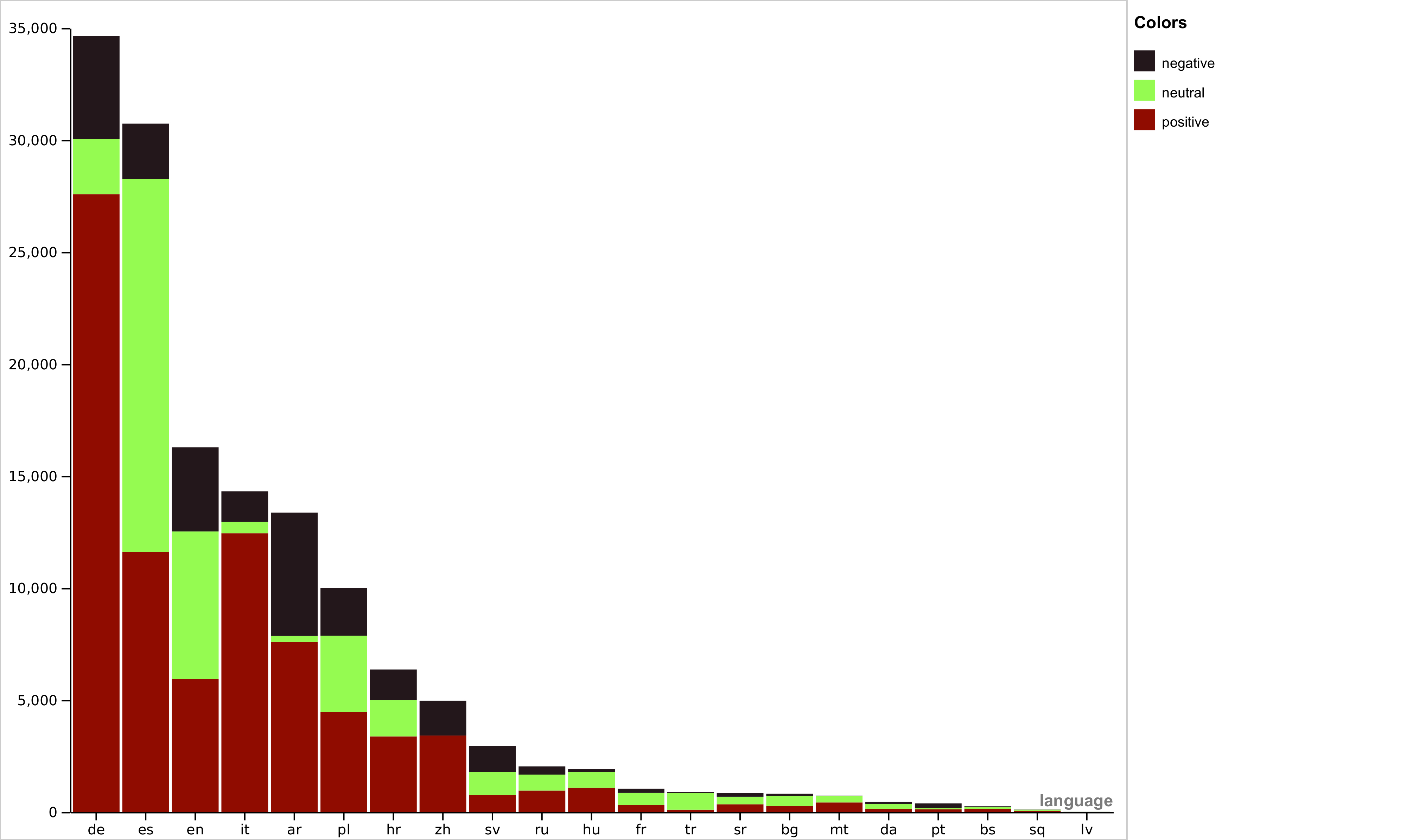} 
\caption{The dataset's distribution across different languages. }
\label{fig.2-dataset-distribution}
\end{center}
\end{figure*}

According to our investigation, numerous datasets are available for conducting both unimodal and multimodal sentiment analysis. The conversion of unimodal datasets, particularly those derived from Twitter, to a multimodal format has been limited. The fundamental hypothesis underlying our utilisation of the unimodal dataset posits that, given its gold annotation, the Twitter dataset can be linked to an image that has not been previously examined or employed in the context of multimodal sentiment classification. Thus, we present our contribution in this field called \textit{M2SA} (Multimodal Multilingual Sentiment Analysis).

\subsection{Data Collection}

An initial step in initiating the enrichment process involves conducting a manual search of pre-existing Twitter sentiment datasets. 
We do not target any other social media datasets to process them uniformly and keep them from a single source. This search is conducted through the utilisation of search engines and data repositories such as HuggingFace Datasets\footnote{\url{https://huggingface.co/datasets}}, European Language Grid\footnote{\url{https://live.european-language-grid.eu}}, and GitHub\footnote{\url{https://github.com/}}. To retrieve the dataset, a search is conducted using specific keywords such as \textbf{twitter sentiment analysis dataset}, \textbf{social media sentiment analysis dataset}, and \textbf{twitter sentiment shared tasks}. Next, the compiled list of datasets undergoes the process of querying tweet information using the Twitter API. The JSON format is used to store the text and images associated with each individual tweet in a dataset.  The initially collected datasets are then subjected to manual checking to exclude tasks unrelated to sentiment analysis. Lastly, label transformations were applied on class labels to convert them from a five-class to a three-class format in cases where they did not initially possess three distinct sentiment categories: positive, negative, and neutral.
The preliminary investigation yielded approximately 100 datasets in multiple languages. However, the final version of our dataset consisted of only 56 distinct datasets, encompassing 21 different languages. This reduction in the number of datasets was primarily due to the absence of tweet IDs linked to the corresponding text in most datasets. Table \ref{dataset-names} presents a comprehensive overview of various languages and their respective datasets that were collected.

\subsection{Preprocessing}
Preprocessing social media texts is imperative due to their inherent informality and noise. The preprocessing steps are delineated as follows:
\begin{itemize}
    \item Removal of all black and white images.
    \item Tweet normalisation for USERs, URLs and HASHTAGS i.e., replace @ElonMusk → <user\_1>…, URLs → <URL\_1>, \#tweet → <hashtag>tweet<hashtag/>
    \item Filtering of tweets with text content less than five characters, not accounting for USER and URL tags.
    \item Deduplication is performed using tweet IDs. 
    \item Checking if the same tweet ID has more than one label assigned and employing a majority vote when needed.
    \item Filtering of tweets with corrupted or no images
    or with images of less than 200 × 200 pixels size.
    \item Checking the language tag in the tweet JSON and see if it matches the target language.
    \item Translation of English tweets for lower-resourced languages using the NLLB\footnote{\url{https://huggingface.co/facebook/nllb-200-3.3B}} machine translation (MT) model.
    
\end{itemize}
The complete preprocessed dataset is structured according to a schema that can be described as follows:

\begin{itemize}
    \item tweetid: unique identifier for the tweet.
    \item normalised-text: text obtained after applying preprocessing steps.
    \item language: the language of the text.
    \item translated-text: text in the target language obtained using the NLLB model. 
    \item image-paths: list of images associated with the tweet.
    \item label: POSITIVE|NEGATIVE|NEUTRAL
\end{itemize}

\subsection{Dataset}

\begin{table}[!ht]
\begin{center}
\begin{tabularx}{\columnwidth}{|c|X|}
\hline
\textbf{Lang} & \textbf{Dataset name}       \\ \hline
ar                & SemEval-2017    \\\hline
ar                & TM-Senti@ar  \\\hline
bg                & Twitter-15@Bulgarian \\\hline
bs                & Twitter-15@Bosnian \\ \hline
da                & AngryTweets                                   \\ \hline
de                & xLiMe@German, Twitter-15@German, TM-Senti@de                               \\ \hline
en                & SemEval-2013-task2, SemEval-2015, SemEval-2016                                \\ \hline
en                & CB COLING2014 vanzo                           \\\hline
en                & CB IJCOL2015 ENG castellucci                  \\\hline
en                & RETWEET                              \\\hline
es                & xLiMe@spanish               \\ \hline
es                & Copres14                    \\ \hline
es                & mavis@tweets                         \\ \hline
es                & Twitter-15@Spanish \\ \hline
es                & JOSA corpus                          \\ \hline
es                & TASS 2018, 2019, 2020                \\ \hline
es                & TASS 2012, 2013, 2014, 2015          \\ \hline
fr                & DEFT 2015                            \\ \hline
hr                & InfoCoV-Senti-Cro-CoV-Twitter        \\ \hline
hr                & Twitter-15@Croatian  \\ \hline
hu                & Twitter-15@Hungarian  \\ \hline
it                & CB IJCOL2015 ITA castellucci           \\ \hline
it                & xLiMe@Italian                 \\ \hline
it                & sentipolc16  \\ \hline
it                & TM-Senti@it                            \\ \hline
lv                & Latvian tweet corpus                   \\\hline
mt                & Malta-Budget-2018, 2019, 2020            \\\hline
pl                & Twitter-15@Polish     \\ \hline
pt                & Twitter-15@Portuguese \\ \hline
pt                & Brazilian tweet@tweets \\\hline
ru                & Twitter-15@Russian   \\\hline
sq                & Twitter-15@Albanian   \\ \hline
sr                & doiserbian@tweet                       \\ \hline
sr                & Twitter-15@Serbian    \\\hline
sv                & Twitter-15@Swedish    \\\hline
tr                & BounTi Turkish       \\\hline
zh                & TM-Senti@zh-ids                        \\\hline
\end{tabularx}
\end{center}
\caption{\label{dataset-names} Languages and their corresponding dataset names}
\end{table}

Figure \ref{fig.2-dataset-distribution} illustrates the comprehensive distribution of datasets across different classes, encompassing 21 languages. The final dataset consists of 143K data points. 

The dataset contains the following languages: Arabic-ar \citep{rosenthal-etal-2017-semeval, Yin2021},
Bulgarian-bg \citep{11356/1054},
Bosnian-bs \citep{11356/1054},
Danish-da \citep{pauli2021danlp},
German-de \citep{11356/1078},
English-en \citep{nakov-etal-2013-semeval,ghosh-etal-2015-semeval,rosenthal-etal-2015-semeval,nakov-etal-2016-semeval,VanzoCB14,castellucci2015context,RETWEET},
Spanish-es \citep{unal_56482,doi:10.1080/21645515.2021.1877597,aguero-torales-etal-2021-logistical,villena2013tass,roman2015tass,vilares2015lys,montejo2016participacion,martinez2018overview,diaz2019overview,garcia2020overview},
French-fr \citep{vukotic2015irisa},
Croatian-hr \citep{app112110442},
Hungarian-hu \citep{11356/1054},
Italian-it \citep{moctezuma2016performance} ,
Maltese-mt \citep{cortis_2021_4650232},
Polish-pl \citep{11356/1054},
Portuguese-pt \citep{Patrick2022-mk},
Russian-ru \citep{11356/1054},
Serbian-sr \citep{articleserbain},
Swedish-sv \citep{11356/1054},
Turkish-tr \citep{mutlu-ozgur-2022-dataset},
Chinese-zh \citep{Yin2021},
Latvian-lv \citep{muischnek2018latvian} and
Albanian-sq \citep{11356/1054}. Languages with more data points, such as German, Spanish, English, Italian, Arabic, and Polish, possess dataset instances exceeding 10,000, whereas other languages exhibit 5,000 or fewer instances of text and images. The dataset exhibits an average token count ranging from 4.25 to 5.94 words, separating each token by a space. One observable pattern is that tweets classified as positive tend to be more likely to be accompanied by images than other categories. The diagram also indicates an imbalance in the datasets across the languages.

\section{Methodology}

\begin{figure*}[!ht]
\begin{center}
\includegraphics[scale=0.5]{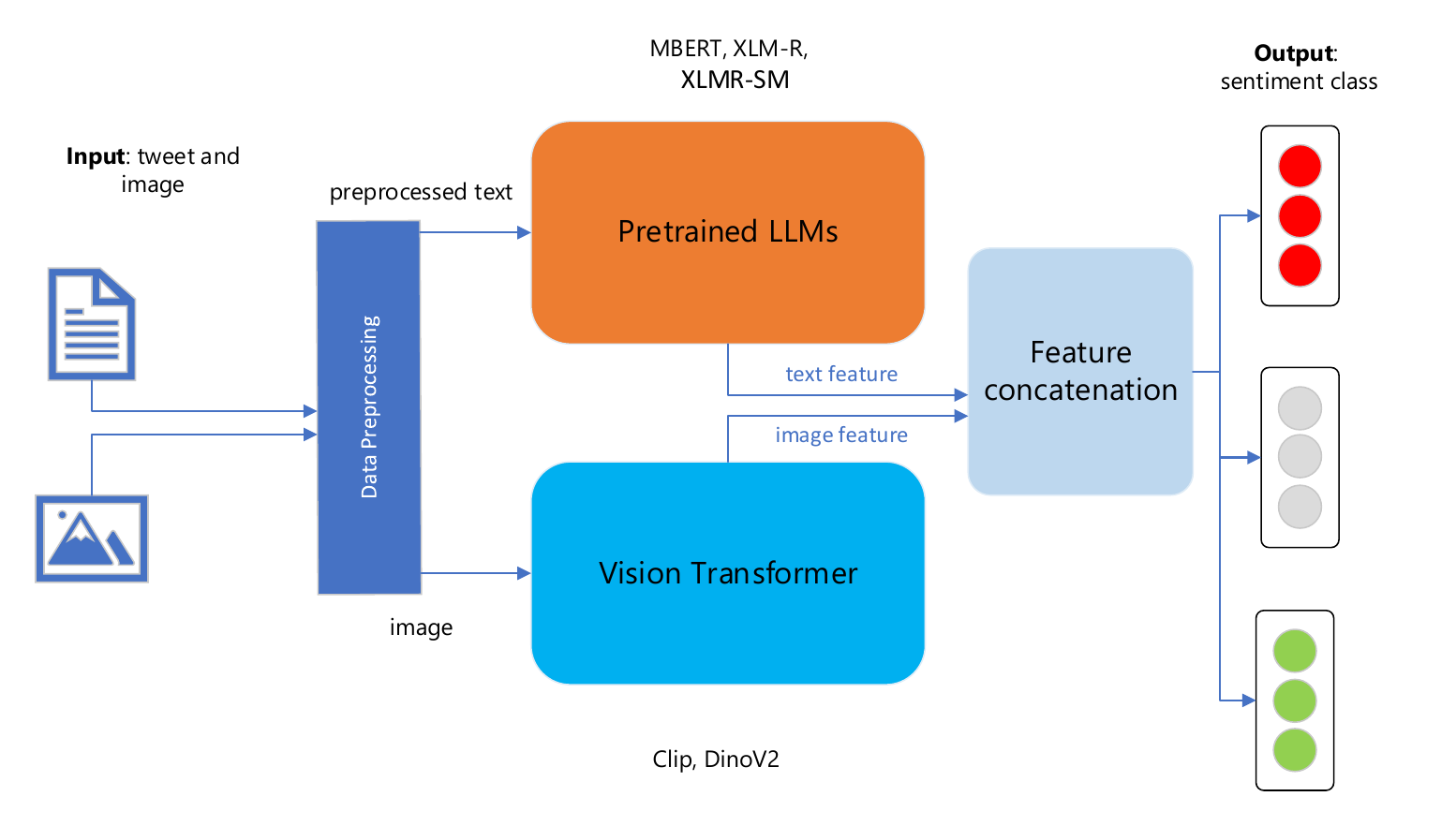} 
\caption{Model architecture of the M2SA.}
\label{fig.abstract}
\end{center}
\end{figure*}

\subsection{Problem Definition}
In the task of unimodal sentiment analysis, the model receives a sequence $X_{m}$ as input, where $m$ represents the length of the sequence. The model then produces a single class as output, which belongs to a closed set consisting of positive, negative, or neutral sentiments. In the context of multimodal sentiment analysis, the model receives input from multiple modalities denoted as $X^{1}_{m}\ldots X^{2}_{n}$, and the output is equivalent to that of unimodal sentiment analysis. The objective of the models is to extract features from the input vectors and acquire the ability to classify sentiment accurately.

The model architecture of the overall sentiment classification system is depicted in Figure~\ref{fig.abstract}.
We examine distinct computational scenarios, encompassing the analysis of textual data alone and the integration of both textual and visual information, to classify the sentiment expressed in tweets. In the context of unimodal textual experiments, the models employed include Multilingual-BERT \citep{devlin-etal-2019-bert}, XLM-RoBERTa \citep{DBLP:journals/corr/abs-1911-02116}, and XLMR-SM, a fine-tuned model specifically designed for sentiment analysis. In the context of multimodal systems, pre-trained vision models (CLIP and DINOv2) are employed as feature extractors. The combined textual and visual features are modelled using a concatenation operator. 

In the context of language processing, datasets pertaining to a specific language are regarded as a cohesive entity. The dataset containing train, validation, and test sets is utilised directly within their respective sets. In this scenario, if there is no distinct set available, we will partition it into train (85\%), test (10\%), and validation (5\%) sets manually.

Given that the text has already undergone processing, no additional processing has been applied to it. The input text undergoes tokenisation, during which it is padded and truncated according to the maximum length supported by the language models. The 'input id' and 'attention mask' are passed into the Language Model (LLM) to extract textual features for each instance in the dataset. Regarding image modality, the images undergo preprocessing through an image preprocessor linked to the corresponding vision models. The image preprocessor's output is subsequently inputted into the vision encoder. The concatenation of the output from the text and vision encoder is subsequently projected onto a linear layer, which is then followed by a softmax layer for the purpose of classification. The models undergo separate fine-tuning processes using a combined dataset comprising samples from multiple languages.
Furthermore, given the limited amount of data available for languages such as Latvian and Albanian, we opt to employ translation techniques to convert the existing text from another dataset into these languages with fewer than 10,000 tweets.
Prior to the translation process, a language detection procedure using an existing model\footnote{papluca/xlm-roberta-base-language-detection} is executed.
This process aims to accurately identify the source text, as the dataset contains text from various languages. Machine translation models rely on providing source and target language codes to perform translations effectively. In the context of the dataset, a language detection process is conducted to classify the instances into their respective source languages. Subsequently, the machine translation pipeline receives each grouped set along with the corresponding source and target language codes. The subsequent subsections discuss the model architectures and pertinent details associated with the models.

\subsection{Text Encoders}
To evaluate the model's efficacy in the absence of visual features, we conducted a standard fine-tuning process utilising the Transformer model's output. Specifically, we employed contextualised sentence embedding, which consists of a 768-element vector. The vector is subsequently fed into a fully connected (FC) layer consisting of three neurons, which is accompanied by a softmax layer for classification. The subsequent text models were employed as encoders for textual data.

\subsubsection{Multilingual-BERT (M-BERT)}
BERT~\citep{DBLP:journals/corr/abs-1810-04805} is a bidirectional transformer pre-trained with the masked language modelling (MLM) and next sentence prediction (NSP) objectives on the top 104 languages with the largest Wikipedia. This model is chosen due to its multilingual nature.

\subsubsection{XLM-RoBERTa (XLM-R)}
The XLM-R model~\citep{conneau-etal-2020-unsupervised} is a large multilingual language model trained on 2.5 TB of filtered Common Crawl data containing 100 languages. The model was trained with the Masked Language Modelling (MLM) objective, with 15\% of the input words masked. The model has been shown to perform really well on downstream tasks when fine-tuned for supervised tasks. XLM-R can understand the input's language solely based on the input IDs without having to use language tensors. This model has been proven to improve the M-BERT scores in various tasks.

\subsubsection{XLM-RoBERTa-Sentiment-Multilingual (XLMR-SM)}
The XLMR-SM model~\citep{dimosthenis-etal-2022-twitter} is a fine-tuned version of XLM-T \citep{barbieri-etal-2022-xlm} on the tweet sentiment multilingual dataset (all), which consists of text from the following languages: Arabic, English, French, German, Hindi, Italian, Portuguese, and Spanish. The XLM-T model has been pre-trained on approximately 198 million multilingual tweets. We introduce this model to study the effect of the presence of sentiment in the pre-trained encoder. Since this model is based on XLM-R, which is trained on tweets and fine-tuned on sentiment datasets, it should perform better at the classification task.

\subsection{Vision Encoders}
The vision encoders divide an image into fixed-size segments and turn them into a sequence that the model can interpret. The encoder analyses the links between these image patches to capture the image's overall meaning, much like transformers do with text. The visual features obtained from the vision models are combined with those obtained from the Transformers text models. The combined output from the encoders is projected into the latest shared space and fine-tuned on a supervised dataset. We employed the following vision encoder models:

\subsubsection{CLIP}
The CLIP model, as described in the paper by~\citet{DBLP:conf/icml/RadfordKHRGASAM21}, is a multimodal framework that combines visual and linguistic information. The CLIP model utilises a transformer architecture, specifically the Vision Transformer (ViT), to extract visual features. Additionally, it employs a causal language model to acquire text features. Consequently, the textual and visual attributes are subsequently mapped onto a latent space with equivalent dimensionality. Computing the similarity score involves calculating the dot product between the projected image and text features.

\subsubsection{DINOv2}
The DINOv2~\citep{oquab2023dinov2} model is a self-supervised learning approach that builds upon the DINO framework proposed by~\citet{caron2021emerging}. The dataset utilised for pre-training purposes is meticulously curated to encompass a diverse range of images sourced from various domains and platforms, including but not limited to natural images, social media images, and product images. This ensures that the acquired features can be applied to diverse practical scenarios.

\section{Experimental Setup}

In this section, we provide details about the implementation and configurations that we used to 
train the model architecture.

\subsection{Implementation}
The neural network's implementation is founded on the PyTorch library. The pre-trained models in the HuggingFace model hub are utilised through direct API calls. All monolingual models employed a batch size of 8 and a learning rate of 3$e^{-5}$. All experiments were conducted using an NVIDIA V100 GPU with a memory capacity of 16GB. The translation module employed the NLLB-200-3.3B model~\citep{costa2022no}, which encompasses all the languages in the dataset that are considered lower-resourced. All multilingual models employed a learning rate of 5$e^{-5}$.

\subsection{Model Configurations}

We used the following configurations to train the model architecture and evaluate the results:
\begin{itemize}
    \item \textbf{Unimodal vs. Multimodal}: First, we experiment with training the unimodal model by using only the text. In another configuration, we train the model using tweets' image and text content. Such a model considers both modalities and predicts the sentiment label jointly.

    \item \textbf{Original data vs. Inclusion of translations}: In one configuration, we used only the extracted tweets as input for the text encoder to train the model. As shown in Figure~\ref{fig.2-dataset-distribution}, not all languages within the curated dataset possess many instances that can be utilised for training purposes. Therefore, the original tweets are machine-translated from English into the target language, and we combine the original text with the translations to train the models for lower-resourced languages.

    \item \textbf{Monolingual vs. Multilingual}: In the monolingual setting, we train separate models for each language using data only from the respective language (either the original data or the addition of translations). In the multilingual setting, the data for all languages are merged, and we train a single model for all languages.
    
\end{itemize}

\section{Evaluation}
In this section, we analyse the outcomes produced by the aforementioned configurations. Additionally, we proceed to analyse the obtained results. The training and early-stopping of the train set are conducted based on the loss observed on the validation set. Final scoring is performed using the test set. The experiments were conducted with five different random seeds (42, 123, 777, 2020, 31337), and the resulting macro F1 scores were reported.

\subsection{Results}
\begin{table*}[!ht]
    \centering
\begin{tabular}{|l|ccccccccc}
\hline
     \multicolumn{1}{|c|}{\textbf{Lang}}                      & \multicolumn{1}{c|}{\textbf{M}}     & \multicolumn{1}{c|}{\textbf{X}}     & \multicolumn{1}{c|}{\textbf{M+C}}   & \multicolumn{1}{c|}{\textbf{X-SM}}  & \multicolumn{1}{c|}{\textbf{M+D}}   & \multicolumn{1}{c|}{\textbf{X-SM+C}}   & \multicolumn{1}{c|}{\textbf{X-SM}} & \multicolumn{1}{c|}{\textbf{M+C}}   & \multicolumn{1}{c|}{\textbf{X-SM+C}} \\ \hline
\multicolumn{1}{|c|}{}     & \multicolumn{6}{c|}{monolingual}                                                                                                                                                   & \multicolumn{3}{c|}{multilingual}                                                            \\ \hline
ar                         & \multicolumn{1}{c|}{57.3}  & \multicolumn{1}{c|}{64.6}  & \multicolumn{1}{c|}{53.6}  & \multicolumn{1}{c|}{66.5}  & \multicolumn{1}{c|}{25.1}  & \multicolumn{1}{c|}{69.1}  & \multicolumn{1}{c|}{41.0 }                     & \multicolumn{1}{c|}{61.3}  & \multicolumn{1}{c|}{\textbf{72.7}}   \\
bg                         & \multicolumn{1}{c|}{51.9}  & \multicolumn{1}{c|}{38.0}  & \multicolumn{1}{c|}{53.7}  & \multicolumn{1}{c|}{\textbf{63.1}}  & \multicolumn{1}{c|}{11.1}  & \multicolumn{1}{c|}{60.5}  & \multicolumn{1}{c|}{53.5 }                     & \multicolumn{1}{c|}{57.8}  & \multicolumn{1}{c|}{60.8}   \\
bs                         & \multicolumn{1}{c|}{62.4} & \multicolumn{1}{c|}{57.0} & \multicolumn{1}{c|}{60.5} & \multicolumn{1}{c|}{64.4} & \multicolumn{1}{c|}{35.4} & \multicolumn{1}{c|}{66.5} & \multicolumn{1}{c|}{ 40.3  }                   & \multicolumn{1}{c|}{63.1}  & \multicolumn{1}{c|}{\textbf{67.9}}  \\
da                         & \multicolumn{1}{c|}{48.8}  & \multicolumn{1}{c|}{34.5} & \multicolumn{1}{c|}{46.9} & \multicolumn{1}{c|}{66.9} & \multicolumn{1}{c|}{21.9} & \multicolumn{1}{c|}{59.1} & \multicolumn{1}{c|}{55.1    }                 & \multicolumn{1}{c|}{57.8} & \multicolumn{1}{c|}{\textbf{75.2}}  \\
de                         & \multicolumn{1}{c|}{68.7} & \multicolumn{1}{c|}{89.1} & \multicolumn{1}{c|}{69.4} & \multicolumn{1}{c|}{90.1} & \multicolumn{1}{c|}{10.7} & \multicolumn{1}{c|}{89.6} & \multicolumn{1}{c|}{56.3   }                  & \multicolumn{1}{c|}{75.3} & \multicolumn{1}{c|}{\textbf{92.9}}  \\
en                         & \multicolumn{1}{c|}{34.1}  & \multicolumn{1}{c|}{18.8} & \multicolumn{1}{c|}{30.4} & \multicolumn{1}{c|}{36.2} & \multicolumn{1}{c|}{6.6}  & \multicolumn{1}{c|}{33.0} &\multicolumn{1}{c|}{ \textbf{64.2}  }                   & \multicolumn{1}{c|}{52.2} & \multicolumn{1}{c|}{53.7}  \\
es                         & \multicolumn{1}{c|}{46.5} & \multicolumn{1}{c|}{22.6} & \multicolumn{1}{c|}{36.9} & \multicolumn{1}{c|}{51.6}  & \multicolumn{1}{c|}{8.0}  & \multicolumn{1}{c|}{46.4}  & \multicolumn{1}{c|}{\textbf{61.4}   }                  & \multicolumn{1}{c|}{49.4} & \multicolumn{1}{c|}{59.6}  \\
fr                         & \multicolumn{1}{c|}{51.1} & \multicolumn{1}{c|}{40.2} & \multicolumn{1}{c|}{50.9} & \multicolumn{1}{c|}{64.5} & \multicolumn{1}{c|}{18.5} & \multicolumn{1}{c|}{64.9} &\multicolumn{1}{c|}{ \textbf{65.8} }                    & \multicolumn{1}{c|}{41.0} & \multicolumn{1}{c|}{51.5}   \\
hr                         & \multicolumn{1}{c|}{58.5} & \multicolumn{1}{c|}{28.7} & \multicolumn{1}{c|}{56.4} & \multicolumn{1}{c|}{\textbf{64.6}} & \multicolumn{1}{c|}{25.7} & \multicolumn{1}{c|}{55.9} & \multicolumn{1}{c|}{40.5    }                 & \multicolumn{1}{c|}{57.7} & \multicolumn{1}{c|}{63.4}  \\
hu                         & \multicolumn{1}{c|}{50.9} & \multicolumn{1}{c|}{43.1} & \multicolumn{1}{c|}{50.5} & \multicolumn{1}{c|}{62.5} & \multicolumn{1}{c|}{17.8} & \multicolumn{1}{c|}{\textbf{66.3}} &\multicolumn{1}{c|}{ 47.3 }                    & \multicolumn{1}{c|}{56.1} & \multicolumn{1}{c|}{63.7}  \\
it                         & \multicolumn{1}{c|}{40.3} & \multicolumn{1}{c|}{29.8} & \multicolumn{1}{c|}{24.0} & \multicolumn{1}{c|}{55.8} & \multicolumn{1}{c|}{4.4}  & \multicolumn{1}{c|}{60.2} &\multicolumn{1}{c|}{ 54.4 }                    & \multicolumn{1}{c|}{56.6} & \multicolumn{1}{c|}{\textbf{63.1}}  \\
mt                         & \multicolumn{1}{c|}{60.3} & \multicolumn{1}{c|}{60.3} & \multicolumn{1}{c|}{60.0} & \multicolumn{1}{c|}{\textbf{68.3}} & \multicolumn{1}{c|}{11.9} & \multicolumn{1}{c|}{62.0} &\multicolumn{1}{c|}{ 35.9 }                    & \multicolumn{1}{c|}{44.0} & \multicolumn{1}{c|}{56.8}  \\
pl                         & \multicolumn{1}{c|}{67.8} & \multicolumn{1}{c|}{45.3} & \multicolumn{1}{c|}{46.2} & \multicolumn{1}{c|}{68.7} & \multicolumn{1}{c|}{12.7} & \multicolumn{1}{c|}{69.5} &\multicolumn{1}{c|}{ 51.2  }                   & \multicolumn{1}{c|}{63.8} & \multicolumn{1}{c|}{\textbf{72.3}}  \\
pt                         & \multicolumn{1}{c|}{67.2}  & \multicolumn{1}{c|}{48.1} & \multicolumn{1}{c|}{51.8} & \multicolumn{1}{c|}{64.3} & \multicolumn{1}{c|}{29.5} & \multicolumn{1}{c|}{\textbf{74.6}} &\multicolumn{1}{c|}{ 48.3 }                    & \multicolumn{1}{c|}{52.8} & \multicolumn{1}{c|}{61.8}   \\
ru                         & \multicolumn{1}{c|}{65.5} & \multicolumn{1}{c|}{43.9} & \multicolumn{1}{c|}{70.6} & \multicolumn{1}{c|}{73.1} & \multicolumn{1}{c|}{27.1} & \multicolumn{1}{c|}{75.3} &\multicolumn{1}{c|}{ 64.9 }                    & \multicolumn{1}{c|}{65.7} & \multicolumn{1}{c|}{\textbf{82.3}}  \\
sr                         & \multicolumn{1}{c|}{42.6} & \multicolumn{1}{c|}{23.4} & \multicolumn{1}{c|}{38.1} & \multicolumn{1}{c|}{49.7}  & \multicolumn{1}{c|}{21.6} & \multicolumn{1}{c|}{43.8} &\multicolumn{1}{c|}{ 48.7}                     & \multicolumn{1}{c|}{49.9} & \multicolumn{1}{c|}{\textbf{65.3}}  \\
sv                         & \multicolumn{1}{c|}{68.2} & \multicolumn{1}{c|}{43.0} & \multicolumn{1}{c|}{59.2} & \multicolumn{1}{c|}{73.1} & \multicolumn{1}{c|}{28.7} & \multicolumn{1}{c|}{73.3} &\multicolumn{1}{c|}{ 54.5    }                  & \multicolumn{1}{c|}{66.0} & \multicolumn{1}{c|}{\textbf{80.2}}  \\
tr                         & \multicolumn{1}{c|}{45.9}  & \multicolumn{1}{c|}{32.1}  & \multicolumn{1}{c|}{44.4} & \multicolumn{1}{c|}{49.6} & \multicolumn{1}{c|}{11.6} & \multicolumn{1}{c|}{49.4} & \multicolumn{1}{c|}{47.9 }                    & \multicolumn{1}{c|}{41.3} & \multicolumn{1}{c|}{47.8}   \\
zh                         & \multicolumn{1}{c|}{57.6} & \multicolumn{1}{c|}{98.9} & \multicolumn{1}{c|}{64.9} & \multicolumn{1}{c|}{99.0} & \multicolumn{1}{c|}{26.3} & \multicolumn{1}{c|}{98.4}  &\multicolumn{1}{c|}{ 43.9 }                    & \multicolumn{1}{c|}{68.7} & \multicolumn{1}{c|}{\textbf{98.4}}  \\
lv                         & \multicolumn{1}{c|}{22.6} & \multicolumn{1}{c|}{19.0} & \multicolumn{1}{c|}{24.8} & \multicolumn{1}{c|}{22.0} & \multicolumn{1}{c|}{21.5} & \multicolumn{1}{c|}{18.1}  & \multicolumn{1}{c|}{\textbf{76.8} }                    & \multicolumn{1}{c|}{52.4} & \multicolumn{1}{c|}{61.6}  \\
sq                         & \multicolumn{1}{c|}{20.7}  & \multicolumn{1}{c|}{20.7}  & \multicolumn{1}{c|}{20.5} & \multicolumn{1}{c|}{20.5} & \multicolumn{1}{c|}{7.8}   & \multicolumn{1}{c|}{20.5} & \multicolumn{1}{c|}{33.7  }                   & \multicolumn{1}{c|}{43.5} & \multicolumn{1}{c|}{45.4}  \\
bg\_mt                     & \multicolumn{1}{c|}{26.1} & \multicolumn{1}{c|}{23.5} & \multicolumn{1}{c|}{25.8} & \multicolumn{1}{c|}{23.5} & \multicolumn{1}{c|}{9.1}  & \multicolumn{1}{c|}{29.4} &         \multicolumn{1}{c|}{}                    & \multicolumn{1}{c|}{}      & \multicolumn{1}{c|}{}       \\
bs\_mt                     & \multicolumn{1}{c|}{17.3} & \multicolumn{1}{c|}{19.0} & \multicolumn{1}{c|}{15.6} & \multicolumn{1}{c|}{18.5} & \multicolumn{1}{c|}{9.1}  & \multicolumn{1}{c|}{20.6} &             \multicolumn{1}{c|}{}              & \multicolumn{1}{c|}{}      & \multicolumn{1}{c|}{}       \\
da\_mt                     & \multicolumn{1}{c|}{20.7} & \multicolumn{1}{c|}{20.7} & \multicolumn{1}{c|}{20.7} & \multicolumn{1}{c|}{24.5} & \multicolumn{1}{c|}{15.0} & \multicolumn{1}{c|}{24.7} &        \multicolumn{1}{c|}{}                     & \multicolumn{1}{c|}{}      & \multicolumn{1}{c|}{}       \\
fr\_mt                     & \multicolumn{1}{c|}{23.1} & \multicolumn{1}{c|}{23.1} & \multicolumn{1}{c|}{23.1} & \multicolumn{1}{c|}{25.8} & \multicolumn{1}{c|}{13.6} & \multicolumn{1}{c|}{23.4} &                           & \multicolumn{1}{|l|}{}      & \multicolumn{1}{c|}{}       \\
hr\_mt                     & \multicolumn{1}{c|}{34.0} & \multicolumn{1}{c|}{25.4} & \multicolumn{1}{c|}{28.9} & \multicolumn{1}{c|}{34.9} & \multicolumn{1}{c|}{16.5} & \multicolumn{1}{c|}{46.9} &                           & \multicolumn{1}{|l|}{}      & \multicolumn{1}{c|}{}       \\
hu\_mt                     & \multicolumn{1}{c|}{28.7} & \multicolumn{1}{c|}{21.2} & \multicolumn{1}{c|}{22.8} & \multicolumn{1}{c|}{28.6} & \multicolumn{1}{c|}{10.3} & \multicolumn{1}{c|}{28}    &                           & \multicolumn{1}{|l|}{}      & \multicolumn{1}{c|}{}       \\
mt\_mt                     & \multicolumn{1}{c|}{30.1} & \multicolumn{1}{c|}{18.6} & \multicolumn{1}{c|}{20.9} & \multicolumn{1}{c|}{43.8} & \multicolumn{1}{c|}{12.0} & \multicolumn{1}{c|}{26.3} &                           & \multicolumn{1}{|l|}{}      & \multicolumn{1}{c|}{}       \\
pt\_mt                     & \multicolumn{1}{c|}{16.4} & \multicolumn{1}{c|}{8.7}   & \multicolumn{1}{c|}{10.5} & \multicolumn{1}{c|}{22.9} & \multicolumn{1}{c|}{23.4} & \multicolumn{1}{c|}{21.9} &                          & \multicolumn{1}{|l|}{}      & \multicolumn{1}{c|}{}       \\
ru\_mt                     & \multicolumn{1}{c|}{41.3} & \multicolumn{1}{c|}{17.8} & \multicolumn{1}{c|}{28.8}  & \multicolumn{1}{c|}{46.9}  & \multicolumn{1}{c|}{23.7} & \multicolumn{1}{c|}{45.6} &                           & \multicolumn{1}{|l|}{}      & \multicolumn{1}{c|}{}       \\
sr\_mt                     & \multicolumn{1}{c|}{18.8} & \multicolumn{1}{c|}{18.8} & \multicolumn{1}{c|}{18.6} & \multicolumn{1}{c|}{25.5} & \multicolumn{1}{c|}{17.8} & \multicolumn{1}{c|}{23.0} &                           & \multicolumn{1}{|l|}{}      & \multicolumn{1}{c|}{}       \\
sv\_mt                     & \multicolumn{1}{c|}{31.7} & \multicolumn{1}{c|}{17.3} & \multicolumn{1}{c|}{24.3} & \multicolumn{1}{c|}{54.6} & \multicolumn{1}{c|}{19.8} & \multicolumn{1}{c|}{34.7} &                           & \multicolumn{1}{|l|}{}      & \multicolumn{1}{c|}{}       \\
tr\_mt                     & \multicolumn{1}{c|}{33.7} & \multicolumn{1}{c|}{30.8} & \multicolumn{1}{c|}{32.5}  & \multicolumn{1}{c|}{31.5} & \multicolumn{1}{c|}{13.8} & \multicolumn{1}{c|}{30.8} &                           & \multicolumn{1}{|l|}{}      & \multicolumn{1}{c|}{}       \\
zh\_mt                     & \multicolumn{1}{c|}{38.2} & \multicolumn{1}{c|}{66.7} & \multicolumn{1}{c|}{38.0} & \multicolumn{1}{c|}{78.1} & \multicolumn{1}{c|}{25.3} & \multicolumn{1}{c|}{85.4}  &                           & \multicolumn{1}{|l|}{}      & \multicolumn{1}{c|}{}\\\hline      
\end{tabular}

    \caption{\label{experimental-results} F1 comparison of models using visual and textual features. M: M-BERT, C: CLIP, X: XLM-Roberta, X-SM: XLM-RoBERTa-Sentiment-Multilingual, D: DINOv2. \{lang\}\_mt: it refers to the model that uses data from original tweets and their translations for that specific lower-resourced language. The value included within a cell containing model headers signifies the model's performance on the test set for the specific language indicated by the lang column. Monolingual training involves the use of data from a single language, whereas multilingual training involves the incorporation of training data from multiple languages. The best result for each language is highlighted in bold.}
    
\end{table*}

The results (F1-score) for the model configurations are given in Table \ref{experimental-results}. 

\textbf{Unimodal vs. Multimodal}: In terms of using textual features to train unimodal models, we can observe that, on average, textual features from XLM-RoBERTa-Sentiment-Multilingual yielded higher F1-scores than Multilingual-BERT or XLM-RoBERTa-base. When we combine both modalities to train multimodal models, we can observe that the combination of XLM-RoBERTa-Sentiment-Multilingual with CLIP (X-SM+C) demonstrated superior performance compared to other multimodal models. The unimodal models of the Bulgarian, Danish, German, Croatian, Maltese, and Chinese languages exhibit superior performance compared to their multimodal counterparts. In contrast, the multimodal model demonstrated superior performance for the remaining higher-resourced languages.

\textbf{Original data vs Inclusion of translated text}: 
In the context of lower-resourced languages, the utilisation of machine-translated instances sourced from higher-resourced languages, such as English, did not yield significant performance improvements. In the context of the Chinese language, including translated instances resulted in a decline in the overall performance. We hypothesise that, in contrast to product and movie reviews, which encompass comprehensive contextual information as a cohesive entity, one single tweet lacks the wider contextual frame. Consequently, the translation of the original language is of lower quality and  results in a modification of the overall meaning. 

\textbf{Monolingual vs Multilingual}: When compared with monolingual models' results, on average, training a single model for all languages yielded the best performance for 17 languages, where results for Croatian, Hungarian, Maltese, Portuguese are higher with monolingual models. It suggests that providing a single model instead of 21 language-specific models is adequate for many languages of interest in this paper. Regarding modality for multilingual configuration, the combination of XLM-RoBERTa-Sentiment-Multilingual with CLIP (X-SM+C) yielded the best performance across many languages. Thus, we can confirm that the model trained with the configuration of multimodal and multilingual achieved the best score for the sentiment analysis of tweets that include both text and image content.

\begin{figure*}[!ht]
\begin{center}
\includegraphics[width=0.9\textwidth]{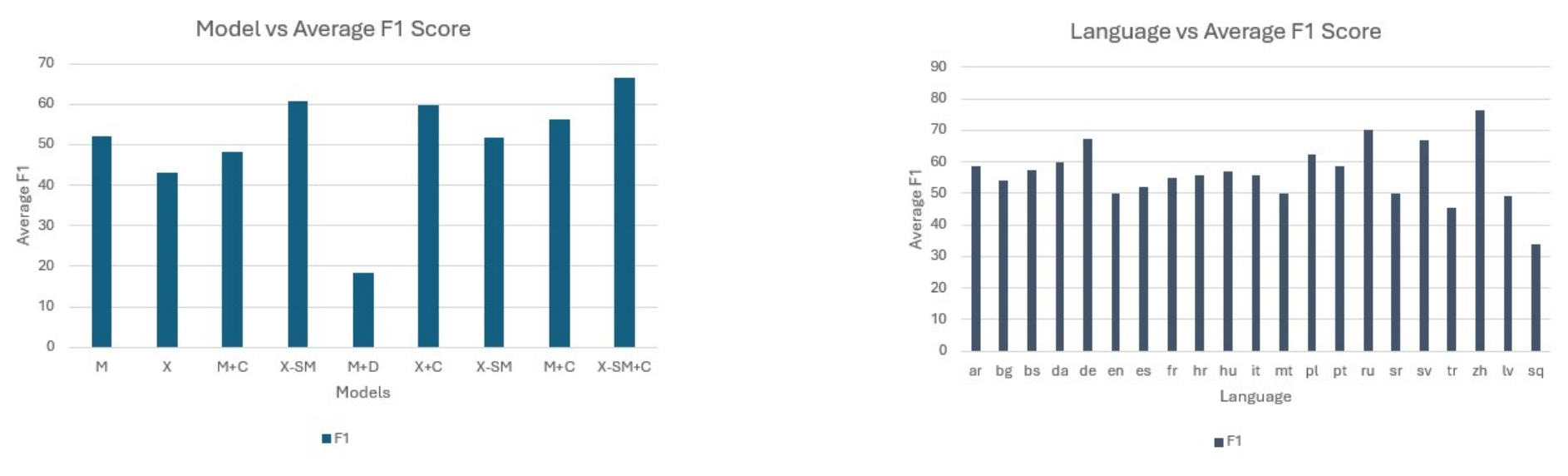} 
\caption{The (left) plot illustrates the averaged F1-score across various models. The (right) plot illustrates the averaged F1-score across languages.}
\label{fig.3}
\end{center}
\end{figure*}

Figure~\ref{fig.3} displays (on right) the average F1-scores for each language and for each combination of pre-trained models (on left). In the first subplot, it is evident that X-SM+C exhibits superior performance across all languages, with XLM-RoBERTa-Sentiment-Multilingual (X-SM) following closely behind. These findings also suggest the significance of pre-trained models, particularly those that are highly specialised or domain-specific in the context of sentiment tasks. In the second subplot, we observe that languages such as Chinese, Russian, Swedish, and German have overall better scores on all the trained models.

\subsection{Error Analysis}
A manual inspection was conducted on the predictions generated by the best performing unimodal and multimodal model. The errors observed in the model can be classified into the following categories:


\textbf{Missing Context}: The tweets exhibited a level of ambiguity that required the application of external knowledge about the world in order to determine the polarity of the messages. Given that tweets often only capture a fragment of a larger conversation and lack the necessary background context, it can be argued that these tweets require additional information beyond the presented text in order to accurately classify their content. The majority of incorrect predictions for unimodal and multimodal can be placed in this category.

\textbf{Disputable}: It is important to note that not all labels present in numerous datasets can be regarded as definitive ground truth, particularly in the case of~\citep{11356/1054}, which has been previously identified as having noise and exhibiting low inter-rater agreement~\citep{rasooli2018cross}. It is our contention that the identification of these instances with noisy labels should be accomplished through the utilisation of established frameworks such as~\citet{NEURIPS_DATASETS_AND_BENCHMARKS2021_f2217062}. This observation suggests that there remains significant potential for enhancement and validates the efficacy of the collaborative assessment of multimodal data. 

\textbf{Figurative Language}: Although the multimodal features help in majority of the case, the models cannot comprehend cases such as sarcasm. In this case, the textual model predicts a neutral class while the multimodal model predicts a positive class, despite the fact that the original class from the dataset is negative.

\begin{figure*}[!ht]
\begin{center}
\includegraphics[width=\textwidth]{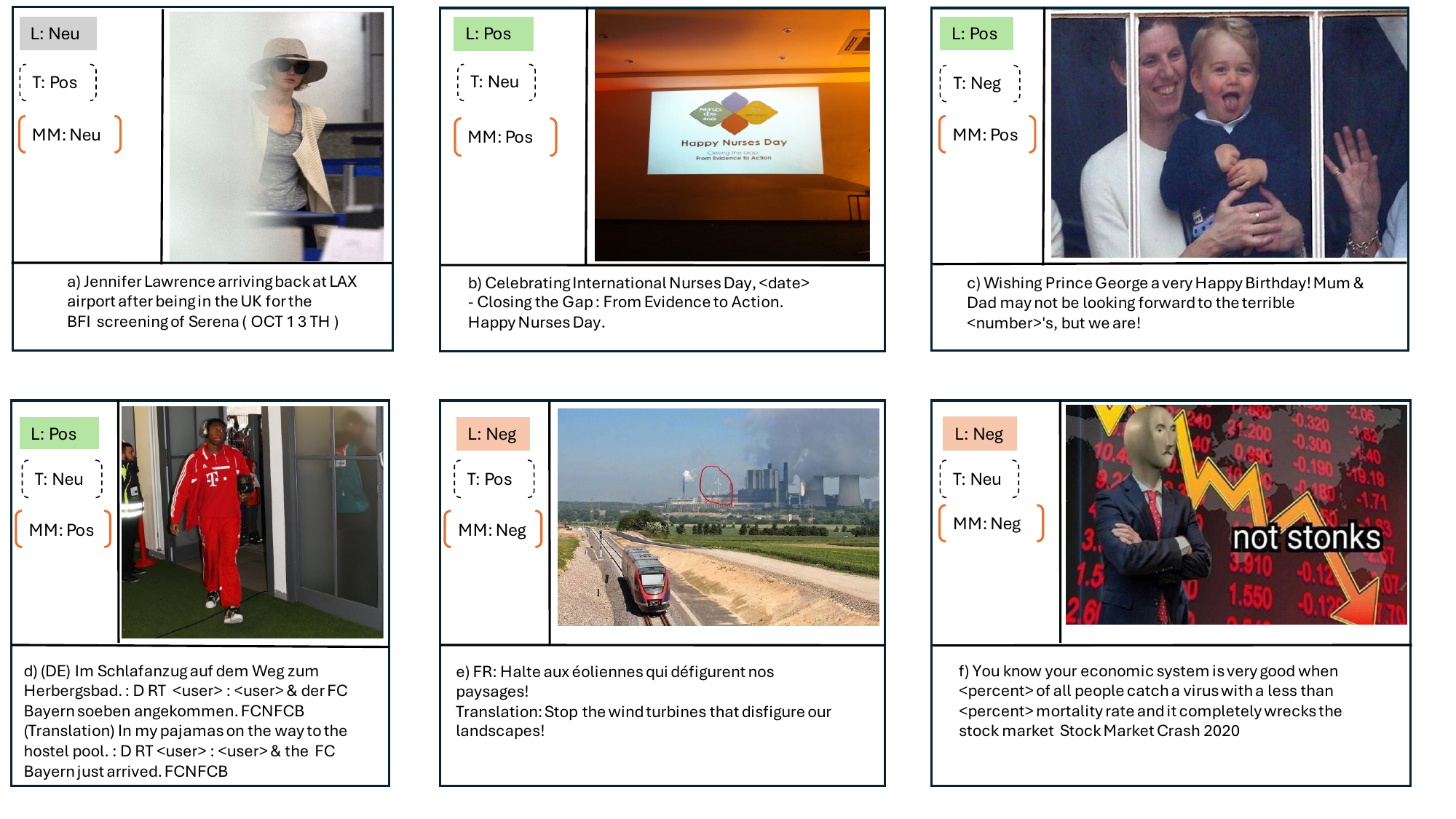} 
\caption{Examples from multilingual multimodal model predicts the correct label and text-only model fail}
\label{fig:examples}
\end{center}
\end{figure*}

In Figure~\ref{fig:examples}, we show a few examples from X-SM+C where the multilingual model predicts the correct label and the unimodal makes an incorrect classification. In the example (c), the tweet contains the text "Wishing Prince George a very Happy Birthday! Mum \& Dad may not be looking forward to the terrible <number>'s, but we are!" is classified as negative by the text model, but the multimodal multimodal multilingual model correctly predicts it as positive.





\section{Conclusion}
This paper presents the model architecture trained on the dataset extracted from various sources for multimodal sentiment classification in a multilingual context. To achieve the objective, this study employed a straightforward methodology to enhance an existing unimodal dataset from Twitter, transforming it into a multimodal one. Numerous models have been trained utilising textual data and a combination of textual and visual modalities. The primary conclusion drawn from this study is that incorporating sentiment knowledge into transformers-based models enhances the accuracy of tweet sentiment classification. The efficacy of the same model settings varies across different languages. Training a single model for all languages multilingual and multimodal data yielded the best performance across many languages. In our prospective endeavours, we intend to utilise tweets devoid of images that underwent filtration during the preprocessing phase. We aim to augment the existing dataset by incorporating additional languages. One potential avenue for advancing research is using translated datasets derived from languages other than the target language.

\section*{Limitations}
The performance of pre-trained models in highly specialised or domain-specific tasks may be limited due to the broad coverage of topics in their training data. The pre-trained models learn from the data they are trained on, which can result in the introduction of any inherent biases in the training data. This bias can affect model outputs, particularly when the data do not represent all demographic, cultural, or social groups. The sentiment datasets used contain a bias towards a particular topic, which was incorporated by the annotators when the datasets were labelled.

\section*{Acknowledgements}
This work was partially funded by the EU Horizon 2020 Research and Innovation Programme under the Marie Sklodowska-Curie grant agreement no. 812997 (CLEOPATRA ITN).
This work was partially funded from the European Union’s Horizon Europe Research and Innovation Programme under Grant Agreement No 101070631 and from the UK Research and Innovation (UKRI) under the UK government's Horizon Europe funding guarantee (Grant No 10039436).

\section{Bibliographical References}\label{sec:reference}
\bibliographystyle{lrec-coling2024-natbib}
\bibliography{lrec-coling2024-example, languageresource}

\end{document}